# Intercausal Independence and Heterogeneous Factorization

**Nevin Lianwen Zhang**
Department of Computer Science
Hong Kong University of Science and Technology
Clear Water Bay, Kowloon, Hong Kong
lzhang@cs.ust.hk

**David Poole**
Department of Computer Science
University of British Columbia
Vancouver, B.C., Canada
poole@cs.ubc.ca

## Abstract

It is well known that conditional independence can be used to factorize a joint probability into a multiplication of conditional probabilities. This paper proposes a constructive definition of intercausal independence, which can be used to further factorize a conditional probability. An inference algorithm is developed, which makes use of both conditional independence and intercausal independence to reduce inference complexity in Bayesian networks.

## 1 INTRODUCTION

In one interpretation of Bayesian networks, arcs are viewed as indication of causality; the parents of a random variable are considered causes that jointly influence the variable (Pearl 1988). The concept *intercausal independence* refers to situations where the mechanism by which a cause influences a variable is independent of the mechanisms by which other causes influence that variable. The noisy OR-gate and noisy adder models (Good 1961, Pearl 1988) are examples of intercausal independence.

Special cases of intercausal independence such as the noisy OR-gate model have been utilized to reduce the complexity of knowledge acquisition (Pearl 1988, Henrion 1987) as well as the complexity of inference (Kim and Pearl 1983). Heckerman (1993) is the first researcher to try to formally define intercausal independence. His definition is temporal in nature. Based on this definition, a graph-theoretic representation of intercausal independence has been proposed.

This paper attempts a constructive definition. Our definition is based on the following intuition about intercausal independence: a number of causes contribute independently to an effect and the total contribution is a combination of the individual contributions. The definition allows us to represent intercausal independence by factorization of conditional probability, in a way similar to that conditional independence can be represented by factorization of joint probability.

The advantages of our factorization-of-conditional-probability representation of intercausal independence over Heckerman's graph-theoretic representation are twofold. Firstly, the symmetric nature of intercausal independence is retained in our representation. Secondly and more importantly, our representation allows one to make full use of intercausal independence to reduce inference complexity.

While Heckerman uses intercausal independencies to alter the topologies of Bayesian networks, we follow Pearl (1988) (section 4.3.2) to exploit intercausal independencies in inference. While Pearl only deals with the case of singly connected networks, we deal with the general case.

The rest of this paper is organized as follows. A constructive definition of intercausal independence is given in Section 2. Section 3 discusses factorization of a joint probability into a multiplication of conditional probabilities, and points out intercausal independence allows one to further factorize conditional probabilities into "even-smaller" factors. The fact that those "even-smaller" factors might be combined by operators other than multiplication leads to the concept of heterogeneous factorization (HF). After some technical preparations (Sections 4 and 5), the formal definition of HF is given in section 6. Section 7 discusses how to sum out variables from an HF. An algorithm for computing marginals from an HF is given in Section 8, which is illustrated through an example in Section 9. Related work is discussed in Section 10.

## 2 CONSTRUCTIVE INTERCAUSAL INDEPENDENCE

This sections gives a constructive definition of intercausal independence. This definition is based on the following intuition: a number of causes $c_1, c_2, \ldots, c_m$ contribute independently to an effect $e$ and the total contribution is a combination of the individual contributions.

Let us begin with an example — the noisy OR-gate model (Good 1961, Pearl 1988, Heckerman 1993). In this model, there is a random binary variable $\xi_i$ in correspondence to each $c_i$, which is also binary. $\xi_i$ depends on $c_i$ and is conditionally independent of any other $\xi_j$ given the $c_i$'s. $e$ is 0 if and only if all the $\xi_i$'s are 0, and is 1 otherwise. In formula, $e = \xi_1 \vee \ldots \vee \xi_m$.

Consider the case when $m=2$ and consider the conditional probability $P(e|c_1, c_2)$. For any value $\beta_i$ of $c_i$ ($i=1, 2$), we have

$$P(e=0|c_1=\beta_1, c_2=\beta_2)$$
$$= P(\xi_1 \vee \xi_2 = 0|c_1=\beta_1, c_2=\beta_2)$$
$$= P(\xi_1=0|c_1=\beta_1)P(\xi_2=0|c_2=\beta_2),$$

and

$$P(e=1|c_1=\beta_1, c_2=\beta_2)$$
$$= P(\xi_1 \vee \xi_2 = 1|c_1=\beta_1, c_2=\beta_2)$$
$$= P(\xi_1=1|c_1=\beta_1)P(\xi_2=0|c_2=\beta_2)$$
$$+ P(\xi_1=0|c_1=\beta_1)P(\xi_2=1|c_2=\beta_2)$$
$$+ P(\xi_1=1|c_1=\beta_1)P(\xi_2=1|c_2=\beta_2).$$

Define $f_1(e=\alpha_1, c_1=\beta_1) =_{def} P(\xi_1=\alpha_1|c_1=\beta_1)$ and define $f_2(e=\alpha_2, c_2=\beta_2) =_{def} P(\xi_2=\alpha_2|c_2=\beta_2)$. We can rewrite the above two equations in a more compact form as follows:

$$P(e=\alpha|c_1=\beta_1, c_2=\beta_2) = \sum_{\alpha_1 \vee \alpha_2 = \alpha} f_1(e=\alpha_1, c_1=\beta_1) f_2(e=\alpha_2, c_2=\beta_2), \quad (1)$$

where $\alpha$, $\alpha_1$, and $\alpha_2$ can be either 0 or 1. This example motivates the following definitions.

Let $e$ be a discrete variable and let $*_e$ be an commutative and associative binary operator over the frame $\Omega_e$ — the set of possible values — of $e$. In the previous example, $*_e$ is the logic OR operator $\vee$. Let $f(e, x_1, \ldots, x_r, y_1, \ldots, y_s)$ and $g(e, x_1, \ldots, x_r, z_1, \ldots, z_t)$ be two functions, where the $y_i$'s are different from the $z_j$'s. Then, the *combination* $f \otimes_e g$ of $f$ and $g$ is defined as follows: for any value $\alpha$ of $e$,

$$f \otimes_e g(e=\alpha, x_1, \ldots, x_r, y_1, \ldots, y_s, z_1, \ldots, z_t)$$
$$=_{def} \sum_{\alpha_1 *_e \alpha_2 = \alpha} [f(e=\alpha_1, x_1, \ldots, x_r, y_1, \ldots, y_s) \times$$
$$g(e=\alpha_2, x_1, \ldots, x_r, z_1, \ldots, z_t)]. \quad (2)$$

We shall refer to $*_e$ as the *base combination operator* and $\otimes_e$ as the $*_e$-*induced combination operator*. We would like to alert the reader that $*_e$ combines values of $e$, while $\otimes_e$ combines functions of $e$. It is easy to see that the induced operator $\otimes_e$ is also commutative and associative.

Here is our constructive definition of intercausal independence. We say that $c_1, \ldots, c_m$ *contribute independently to* $e$ or $e$ *receives contributions independently from* $c_1, \ldots, c_m$ if there exists a commutative and associative binary operator $*_e$ over the frame of $e$ and real-valued non-negative functions $f_1(e, c_1), \ldots, f_m(e, c_m)$ such that

$$P(e|c_1, \ldots, c_m) = f_1(e, c_1) \otimes_e \ldots \otimes_e f_m(e, c_m), \quad (3)$$

where $\otimes_e$ is the $*_e$-induced combination operator. The right hand of the equation makes sense because $\otimes_e$ is commutative and associative. When $c_1, \ldots, c_m$ contribute independently to $e$, we call $e$ a *bastard variable*[1]. A non-bastard variable is said to be *normal*. We also say that $f_i(e, c_i)$ is *(a description of) the contribution by $c_i$ to $e$*.

Intuitively, the base combination operator (e.g. $\vee$) determines how contributions from different sources are combined, while the induced combination operator is the reflection of the base operator at the level of conditional probability.

Because of equation (1), the noisy-OR gate model is an example of constructive intercausal independence, with the logic OR $\vee$ as the base combination operator.

As another example, consider the noisy adder model (Heckerman 1993). In this model, there is a random variable $\xi_i$ in correspondence to each $c_i$; $\xi_i$ depends on $c_i$ and is conditionally independent of any other $\xi_j$ given the $c_i$'s. The $\xi_i$'s are combined by the addition operator "+" to result in $e$, i.e. $e = \xi_1 + \ldots + \xi_m$.

To see that $e$ is a bastard variable in this model, let the base combination operator $*_e$ be simply "+" and let the description of individual contribution $f_i(e, c_i)$ be as follows: for any value $\alpha$ of $e$ and any value $\beta$ of $c_i$,

$$f_i(e=\alpha, c_i=\beta) =_{def} P(\xi_i=\alpha|c_i=\beta).$$

Then it is easy to verify that equation (3) is satisfied.

It is interesting to notice the similarity between equation (3) and the following property of conditional independence: if a variable $x$ is independent of another variable $z$ given a third variable $y$, then there exist non-negative functions $f(x, y)$ and $g(y, z)$ such that the joint probability $P(x, y, z)$ is given by

$$P(x, y, z) = f(x, y) g(y, z). \quad (4)$$

---

[1]Those who are familiar with clique tree propagation may remember that the first thing to do in constructing a clique tree from a Bayesian network is to "marry" the parents of each node (variable) (Lauritzen and Spiegehalter 1988). As implies by the word "bastard", the parents of a bastard node will not be married. This is because the conditional probability of a bastard node is factorized into a bunch of factors, each involving only one parent.

In (4) conditional independence allows us to factorize a joint probability into factors that involve less variables, while in (3) intercausal independence allows us to factorize a conditional probability into a bunch of factors that involve less variables. The only difference lies in the way the factors are combined.

Conditional independence has been used to reduce inference complexity in Bayesian networks. The rest of this paper investigates how to use intercausal independence for the same purpose.

## 3  FACTORIZATION OF JOINT PROBABILITIES

This section discusses factorization of joint probabilities and introduces the concept of heterogeneous factorization (HF).

A fundamental assumption under the theory of probabilistic reasoning is that a joint probability is adequate for capturing experts' knowledge and beliefs relevant to a reasoning-under-uncertainty task. Factorization and Bayesian networks come into play because joint probability is difficult, if not impossible, to directly assess, store, and reason with.

Let $P(x_1, x_2, \ldots, x_n)$ be a joint probability over variables $x_1, x_2, \ldots, x_n$. By the chain rule of probabilities, we have

$$P(x_1, x_2, \ldots, x_n) = P(x_1) P(x_2|x_1) \ldots P(x_n|x_1, \ldots, x_{n-1}). \quad (5)$$

For any $i$, there might be a subset $\pi_i \subseteq \{x_1, \ldots, x_{i-1}\}$ such that $x_i$ is conditionally independent of all the other variables in $\{x_1, \ldots, x_{i-1}\}$ given the variables in $\pi_i$, i.e $P(x_i|x_1, \ldots, x_{i-1}) = P(x_i|\pi_i)$. Equation (5) can hence be rewritten as

$$P(x_1, x_2, \ldots, x_n) = \prod_{i=1}^{n} P(x_i|\pi_i). \quad (6)$$

Equation (6) factorizes the joint probability $P(x_1, x_2, \ldots, x_n)$ into a multiplication of factors $P(x_i|\pi_i)$. While the joint probability involves all the $n$ variables, the factors usually involves less than $n$ variables. This fact implies savings in assessing, storing, and reasoning with probabilities.

A *Bayesian network* is constructed from the factorization as follows: construct a directed graph with nodes $x_1, x_2, \ldots, x_n$ such that there is an arc from $x_j$ to $x_i$ if and only if $x_j \in \pi_i$, and associate the conditional probability $P(x_i|\pi_i)$ with the node $x_i$. $P(x_1, \ldots, x_n)$ is said to be the *joint probability* of the Bayesian network so constructed. Also nodes in $\pi_i$ are called *parents* of $x_i$.

The above factorization is *homogeneous* in the sense that all the factors are combined in the same way, i.e by multiplication.

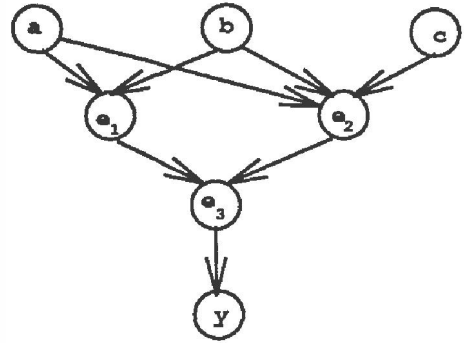

Figure 1: A Bayesian network, where $e_1$ and $e_2$ receive contribution independently from their respective parents.

Let $x_{i1}, \ldots, x_{im_i}$ be the parents of $x_i$. If $x_i$ is a bastard variable with base combination operator $*_i$, then the conditional probability $P(x_i|\pi_i)$ can be further factorized by

$$P(x_i|\pi_i) = f_{i1}(x_i, x_{i1}) \otimes_i \ldots \otimes_i f_{im_i}(x_i, x_{im_i}), \quad (7)$$

where $\otimes_i$ is the $*_i$-induced combination operator. The fact that $\otimes_i$ might be other than multiplication leads to the concept of heterogeneous factorization (HF). The word heterogeneous reflects the fact that different factors might be combined in different manners.

As an example, consider the Bayesian network in Figure 1. The network indicates that $P(a, b, c, e_1, e_2, e_3, y)$ can be factorized into a multiplication of $P(a)$, $P(b)$, $P(c)$, $P(e_1|a, b)$, $P(e_2|a, b, c)$, $P(e_3|e_1, e_2)$, and $P(y|e_3)$.

Now if the $e_i$'s are bastard variables, then there exist base combination operators $*_i$ (i=1, 2, 3) such that the conditional probabilities of the $e_i$'s can be further factorized as follows:

$$\begin{aligned} P(e_1|a, b) &= f_{11}(e_1, a) \otimes_1 f_{12}(e_1, b) \\ P(e_2|a, b, c) &= f_{21}(e_2, a) \otimes_2 f_{22}(e_2, b) \otimes_2 f_{23}(e_2, c) \\ P(e_3|e_1, e_2) &= f_{31}(e_3, e_1) \otimes_3 f_{32}(e_3, e_1) \end{aligned}$$

where $f_{11}(e_1, a)$, for instance, denotes the contribution by $a$ to $e_1$, and where the $\otimes_i$'s are the combination operators respectively induced by the $*_i$'s.

The factorization of $P(a, b, c, e_1, e_2, e_3, y)$ into the factors: $P(a)$, $P(b)$, $P(c)$, $P(y|e_3)$, $f_{11}(e_1, a)$, $f_{12}(e_1, b)$, $f_{21}(e_2, a)$, $f_{22}(e_2, b)$, $f_{23}(e_2, c)$, $f_{31}(e_3, e_1)$, and $f_{32}(e_3, e_2)$ is called the *HF in correspondence to the Bayesian network in Figure 1*. We shall call the $f_{ij}$'s *heterogeneous factors* since they might be combined by operators other than multiplication. On the other hand, we shall say that the factors $P(a)$, $P(b)$, $P(c)$, and $P(y|e_3)$ are *normal*.

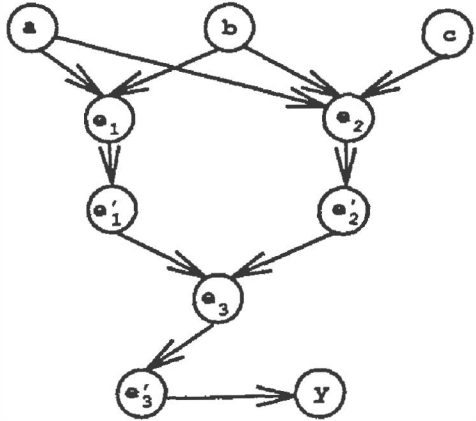

Figure 2: The Bayesian network in Figure 1 after the deputation of bastard nodes.

## 4 DEPUTATION OF BASTARD NODES

Consider the heterogeneous factor $f_{31}(e_3, e_1)$ from the previous example. It contains two bastard variables $e_1$ to $e_3$. As we shall see later, it is desirable for every heterogeneous factor to contain at most one bastard variable. The concept of deputation is introduced to guarantee this.

Let $e$ be a bastard node in a Bayesian network. The *deputation of* $e$ is the following operation: make a copy $e'$ of $e$, make the children of $e$ to be children of $e'$, make $e'$ a child of $e$, and set the conditional probability $P(e'|e)$ as follows:

$$P(e'|e) = \begin{cases} 1 & \text{if } e = e' \\ 0 & \text{otherwise} \end{cases} \quad (8)$$

We shall call $e'$ the *deputy* of $e$. We shall also call $P(e|e')$ the *deputing function*, and rewrite it as $I(e, e')$ since $P(e|e')$ ensures that $e$ and $e'$ be the same.

The Bayesian network in Figure 1 becomes the one in Figure 2 after the deputation of all the bastard nodes. We shall call the latter a *a deputation Bayesian network*.

**Proposition 1** *Let $\mathcal{N}$ be a Bayesian network, and let $\mathcal{N}'$ is the Bayesian network obtained from $\mathcal{N}$ by the deputation of all bastard nodes. Then the joint probability of $\mathcal{N}$ can be obtained from that of $\mathcal{N}'$ by summing out all the deputy variables.* □

In Figure 1, we have the heterogeneous factors $f_{31}(e_3, e_1)$ and $f_{32}(e_3, e_2)$, which involves two bastard variables. This may cause confusions and is undersirable for other reasons, as we shall see soon. After deputation, each heterogeneous factor involves only one bastard variable. As a matter of fact, $f_{31}(e_3, e_1)$ and $f_{32}(e_3, e_2)$ have become $f_{31}(e_3, e'_1)$ and $f_{32}(e_3, e'_2)$.

To prevent $I(e_1, e'_1)$ from being mistaken to be the contribution by $e'_1$ to $e_1$, we shall always make it explicit that $I(e_1, e'_1)$ is a normal factor, not a heterogeneous factor.

## 5 COMBINING FACTORS THAT INVOLVE MORE THAN ONE BASTARD VARIABLE

Even though deputation guarantees that every heterogeneous factor involves only one bastard variable at the beginning, inference may give rise to factors that involve more than one bastard variable. In Figure 2, for instance, summing out the variable $a$ results in a factor that involves both $e_1$ and $e_2$. This section introduces an operator for combining such factors.

Suppose $e_1, \ldots, e_k$ are bastard variables with base combination operator $*_1, \ldots, *_k$. Let $f(e_1, \ldots, e_k, x_1, \ldots, x_r, y_1, \ldots, y_s)$ and $g(e_1, \ldots, e_k, x_1, \ldots, x_r, z_1, \ldots, z_t)$ be two functions, where the $x_i$'s are normal variables and the $y_j$'s are different from the $z_l$'s (they can be bastard as well as normal variables). Then, the *combination* $f \otimes g$ of $f$ and $g$ is defined as follows: for any particular value $\alpha_i$ of $e_i$,

$$\begin{aligned} & f \otimes g(e_1 = \alpha_1, \ldots, e_k = \alpha_k, x_1, \ldots, x_r, \\ & \qquad y_1, \ldots, y_s, z_1, \ldots, z_t) \\ & =_{def} \sum_{\alpha_{11} *_1 \alpha_{12} = \alpha_1} \cdots \sum_{\alpha_{k1} *_k \alpha_{k2} = \alpha_2} \\ & [f(e_1 = \alpha_{11}, \ldots, e_k = \alpha_{k1}, x_1, \ldots, x_r, y_1, \ldots, y_s) \times \\ & \; g(e_1 = \alpha_{12}, \ldots, e_k = \alpha_{k2}, x_1, \ldots, x_r, z_1, \ldots, z_t)]. \end{aligned} \quad (9)$$

A few notes are in order. First, fixing a list of bastard variables and their base combination operators, one can use the operator $\otimes$ to combined two arbitrary functions. In the following, we shall always work implicitly with a fixed list of bastard variables, and we shall refer to $\otimes$ as the *general combination operator*.

Second, when $k = 1$ this definition reduces to equation (2).

Third, since the base combination operators are commutative and associative, the operator $\otimes$ is also commutative and associative.

Fourth, when $k = 0$, $f \otimes g$ is simply the multiplication of $f$ and $g$.

### 5.1 Combining all the Heterogeneous Factors in a Bayesian networks

Equipped with the general combination operator $\otimes$, we now consider combining all the heterogeneous factors of the Bayesian network in Figure 2. Because of the third note above, we can combine them in any order. Let us first combine $f_{11}(e_1, a)$ with $f_{12}(e_2, b)$, $f_{21}(e_2, a)$ with $f_{22}(e_2, b)$ and $f_{23}(e_2, c)$, and $f_{31}(e_3, e'_1)$ with $f_{32}(e_3, e'_2)$. Because of the second note, we have

$$f_{11} \otimes f_{12}(e_1, a, b) = P(e_1|a, b),$$
$$f_{21} \otimes f_{22} \otimes f_{23}(e_2, a, b, c) = P(e_2|a, b, c),$$
$$f_{31} \otimes f_{32}(e_3, e'_2, e'_1) = P(e_3|e'_2, e'_1).$$

We now combine the resulting conditional probabilities. Because of the fourth note, the combination of $P(e_1|a, b)$, $P(e_2|a, b, c)$, and $P(e_3|e'_1, e'_2)$ is their multiplication. So, the combination of all the heterogeneous factors of the Bayesian network in Figure 2 is simply the multiplication of the conditional probabilities of all the bastard variables. This is true in general.

**Proposition 2** *In a deputation Bayesian network, the multiplication of the conditional probabilities of all the bastard variables is the same as the result of combining of all the heterogeneous factors.* □

Note that in Figure 1, since $f_{31}(e_3, e_1)$ and $f_{32}(e_3, e_2)$ involve two bastard variables, the combination $f_{11}(e_1, a) \otimes \ldots \otimes f_{23}(e_2, c) \otimes f_{31}(e_3, e_1) \otimes f_{32}(e_3, e_2)$ would not the same as the multiplication of the conditional probabilities of the bastard variables.

This is why we need deputation; deputation allows us to combine the heterogeneous factors by a single combination operator $\otimes$, which opens up the possibility of combining the heterogeneous factors in any order we choose. This flexibility turns out to be the key to the method of utilizing intercausal independence we are proposing in this paper.

## 6 HETEROGENEOUS FACTORIZATION

We now formally define the concept of heterogeneous factorization. Let $X$ be a set of discrete variables. A *heterogeneous factorization (HF)* $\mathcal{F}$ over $X$ consists of

1. A list $e_1, \ldots, e_m$ of variables in $X$ that are said to be *bastard variables*. Associated with each bastard variable $e_i$ is a base combination operator $*_i$, which is commutative and associative,

2. A set $\mathcal{F}_0$ of *heterogeneous factors*, and

3. A set $\mathcal{F}_1$ of *normal factors*.

We shall write an HF as a quadruplet $\mathcal{F} = (X, \{(e_1, *_1), \ldots, (e_m, *_m)\}, \mathcal{F}_0, \mathcal{F}_1)$. Variables that are not bastard are called *normal*.

In an HF, the *combination* $F_0$ of all the heterogeneous factors is given by

$$F_0 =_{def} \otimes_{f \in \mathcal{F}_0} f. \qquad (10)$$

The *joint* $F(X)$ of an HF is the multiplication of $F_0$ and all the normal factors. In formula

$$F =_{def} (\otimes_{f \in \mathcal{F}_0} f) \prod_{g \in \mathcal{F}_1} g. \qquad (11)$$

In the following, we shall also say that the $\mathcal{F}$ is an HF of the function $F(X)$.

### 6.1 HF's in Correspondence to Deputation Bayesian Networks

Suppose $\mathcal{N}$ is a deputation Bayesian network. Suppose $\mathcal{F}$ is the HF that corresponds to $\mathcal{N}$. $\mathcal{F}$ has two interesting properties.

First, according to Proposition 2 the combination of all the heterogeneous factors is the multiplication of the conditional probabilities of all the bastard variables. Thus, the joint of $\mathcal{F}$ is simply the joint probability of $\mathcal{N}$.

**Proposition 3** *The joint of the HF that corresponds to a deputation Bayesian network $\mathcal{N}$ is the same as the joint probability of $\mathcal{N}$.*

To reveal the second interesting property, let us first define the concept of tidness. An HF is *tidy* if for each bastard variable $e$, there exists at most one normal factor that involves $e$. Moreover, this factor, if exists, involves only one other variable in addition to $e$ itself.

An HF that corresponds to a deputation Bayesian network is tidy. For each bastard variable $e$, $I(e, e')$ is the only one normal factor that involves $e$, and this factor involves only one other variable, namely $e'$.

Tidy HF's do not have to be in correspondence to a deputation Bayesian network. As a matter of fact, we shall start with a tidy HF that corresponds to a deputation Bayesian network, and then sum out variables from the HF. We shall sum out variables in such a way such that the tidness is retained. Even though the HF we start out with corresponds to a deputation Bayesian network, after summing out some variables, the resulting tidy HF might no longer correspond to any deputation Bayesian network.

However, we shall continue to use the terms deputy variable and deputing function.

## 7 SUMMING OUT VARIABLES FROM TIDY HF'S

Let $F(X)$ be a function. Suppose $A$ is a subset of $X$. The *projection* $F(A)$ of $F(X)$ onto $A$ is obtained from $F(X)$ by summing out all the variables in $X - A$. In formula

$$F(A) =_{def} \sum_{X - A} F(X). \qquad (12)$$

When $F(X)$ is a joint probability, $F(A)$ is a *marginal probability*.

Summing variables out directly from $F(X)$ usually require too many additions. Suppose $X$ contains $n$ variables and suppose all variables are binary. One needs to perform $2^n - 1$ additions to sum out one variable.

A better idea is to sum out variables from an factorization of $F(X)$ if there is one. This section investigates how to sum out variables from tidy HF's. The following two lemmas are of fundamental importance, and they readily follow the definition of the general combination operator $\otimes$.

**Lemma 1** *Both multiplication and $\otimes$ are distributive w.r.t summation. More specifically, suppose $f$ and $g$ are two functions and variable $x$ appears in $f$ and not in $g$. Then*

1. $\sum_x (fg) = (\sum_x f)g$, and
2. $\sum_x (f \otimes g) = (\sum_x f) \otimes g$.

□

The following lemma spells out two conditions under which multiplication and $\otimes$ are associative with each other.

**Lemma 2** *Let $f$ and $g$ be two functions.*

1. *If $h$ is a function that involves no bastard variables, then*

$$h\{f \otimes g\} = \{hf\} \otimes g. \quad (13)$$

2. *If $h$ is a function such that all the bastard variables in $h$ appear only in $f$ and not in $g$, then*

$$h\{f \otimes g\} = \{hf\} \otimes g. \quad (14)$$

□

We now proceed to consider the problem of summing out variables from a tidy HF in such a way that the tidness is retained. First of all the following proposition deals with the case when the variable to be summed out appears in only one factor.

**Proposition 4** *Let $\mathcal{F}$ be an HF of $F(X)$ and is tidy. Suppose $z$ is a variable that appears only in one factor $f(A)$, normal or heterogeneous. Define $h$*

$$h(A - \{z\}) =_{def} \sum_z f(A).$$

*Let $\mathcal{F}'$ be the HF obtained from $\mathcal{F}$ by replacing $f$ with $h$[2]. Then, $\mathcal{F}'$ is a HF of $F(X-\{z\})$ — the projection of $F(X)$ onto $X-\{z\}$. Moreover if $z$ is not a deputy variable, then $\mathcal{F}'$ remains tidy.*

**Proof**: The first part of proposition follows from Lemma 1.

For the second part, since $z$ is not a deputy variable, it can be either a non-deputy normal variable or a bastard variable. When $z$ is a non-deputy normal variable,

---

[2] The factor $h$ is heterogeneous or normal if and only if $f$ is.

summing out $z$ does not affect the deputing functions. Therefore, $\mathcal{F}'$ remains tidy.

When $z$ is a bastard variable, summing out $z$ will not affect the deputing functions of any other bastard variables. Therefore, $\mathcal{F}'$ also remains tidy. □

In general, a variable can appear in more than one normal and heterogeneous factors. The next proposition reduces the general case to the case where the variable appear in at most two factors, one normal and one heterogeneous.

**Proposition 5** *Let $\mathcal{F}$ be an HF of $F(X)$, and let $z$ be a variable in $X$. Let $f_1, \ldots, f_m$ be all the heterogeneous factors that involve $z$ and let $g_1, \ldots, g_n$ be all the normal factors that involve $z$. Define*

$$f =_{def} \otimes_{i=1}^m f_i,$$

$$g =_{def} \prod_{j=1}^n g_j.$$

*Let $\mathcal{F}'$ be the HF obtained from $\mathcal{F}$ by removing the $f_i$'s and the $g_j$'s, and by adding a new heterogeneous factor $f$ and a new normal factor $g$. Then*

1. *$\mathcal{F}'$ is also an HF of $F(X)$, and $f$ and $g$ are the only two factors that involve $z$. In particular, when either $m=0$ or $n=0$, there is only one factor in $\mathcal{F}'$ that involves $z$.*

2. *If $z$ is not a deputy variable, then when $\mathcal{F}$ is tidy, so is $\mathcal{F}'$.*

**Proof**: The first part of the proposition follows from the commutativity and associativity of multiplication and of the general combination operator $\otimes$.

For the second part, since $z$ is not a deputy variable, it can either be a non-deputy normal variable or a bastard variable. When $z$ is a non-deputy normal variables, the operations performed by the proposition do not affect the deputing functions. Thus, $\mathcal{F}'$ remains tidy.

When $z$ is a bastard variable, the deputing functions are not affect either. Because for each bastard variable $e$, its deputing functions is the only normal factor that involves $e$. So, $\mathcal{F}'$ also remains tidy. □.

The following proposition merges a normal factor into a heterogeneous factor.

**Proposition 6** *Let $\mathcal{F}$ be an HF of $F(X)$ and is tidy. Suppose $z$ is a variable that appears in only one normal factor $g$ and only one heterogeneous factor $f$. Define $h$ by*

$$h =_{def} fg.$$

*Let $\mathcal{F}'$ be the HF obtained from $\mathcal{F}$ by removing $g$ and $f$, and by adding a heterogeneous factor $h$. If $z$ is not*

*a deputy variable, then the joint of $\mathcal{F}'$ is also $F(X)$ and $\mathcal{F}'$ is tidy. Moreover, $h$ is only one factor in $\mathcal{F}'$ that involves $z$.*

**Proof**: We first consider the case when $z$ is a non-deputy normal variable. Because the tidness of $\mathcal{F}$, $g$ involves no bastard variables. According to Lemmas 2 (1), the joint of $\mathcal{F}'$ is also $F$.

Since $g$ is not a deputing function, the operation of combining $f$ and $g$ into one factor does not affect the deputing functions. Hence, $\mathcal{F}'$ remains tidy.

Let us now consider the case when $z$ is a bastard variable. Since $\mathcal{F}$ is tidy, $g$ must be the deputing function of $z$. Since $f$ is the only heterogeneous factor that involves $z$, all other heterogeneous factors do not involve $z$. According Lemma 2 (2), the joint of $\mathcal{F}'$ is also $F$.

After combining $f$ and $g$ into a heterogeneous factor, there is no normal factor that involve $z$. Also, the deputing functions of the other bastard variables are not affected. Hence, $\mathcal{F}'$ remains tidy. □

The above three propositions allow us to sum out, from a tidy HF, bastard variables and non-deputy normal variables. You may ask: how about deputy variables? As it turns out, after summing out a bastard variable $e$, its deputy $e'$ becomes a non-deputy normal variable. So, we can also sum out deputy variables; we just have to make sure to sum out a deputy variable *after* the corresponding bastard variable has been summed out.

It is possible to intuitively understand why a deputy variable $e'$ needs to be summed out after the corresponding bastard variable $e$. As a matter of fact, summing out $e'$ before $e$ is the inverse of the deputation of $e$. But we have shown at the end the Section 5 that deputation is necessary.

## 8   AN ALGORITHM

This section presents an algorithm for computing projections of a function $F(X)$ by summing variables from a tidy HF of $F(X)$. Because of Proposition 3, the algorithm can be used to compute marginal probabilities, and hence posterior probabilities, in Bayesian networks.

To sum out the variables in $X-A$, an ordering needs to be specified (Lauritzen and Spiegelhalter 1988). In the literature, such an ordering is called an *elimination ordering*, which can be found by heuristics such as the maximum cardinality search (Tarjan and Yannakakis 1984) or the maximal intersection search (Zhang 1993).

At the end of the last section, we said that a deputy variable should be summed out only after the corresponding bastard variable has been summed out. If $e$ is a bastard variable in $A$, what should we do with its deputy variable $e'$?

The paper is concerned with intercausal independence in Bayesian networks. To this end, we need only consider deputing functions $I(e, e')$ such that $I(e, e') = 1$ if $e = e'$ and $I(e, e') = 0$ otherwise. Let us say such deputing functions are *identifying*. Since for any function $f(e, e', x_1, \ldots, x_n)$,

$$\sum_{e'} I(e, e') f(e, e', x_1, \ldots, x_n) = f(e, e, x_1, \ldots, x_n),$$

we can handle the deputies of bastard variables in $A$ as follows: wait till after all the other variables outside $A$ have been summed out and all the heterogeneous factors have been combined, then simply remove all the deputing functions, replace each occurrence of a deputy variable with the corresponding bastard variable. This operation can be viewed as the inverse of deputation.

Procedure PROJECTION($\mathcal{F}, A, \rho$)

- Input:
  1. $\mathcal{F}$ — A tidy HF of a certain function $F(X)$ such that all the deputing functions are identifying,
  2. $A$ — A subset of $X$,
  3. $\rho$ — An elimination ordering consisting all the variables other than the variables $A$ and their deputies. In $\rho$, a deputy variable $e'_i$ comes right after the corresponding bastard variable $e_i$.
- Output: $F(A)$ — The projection of $F$ onto $A$.

1. If $\rho$ is empty, combine all the heterogeneous factors by using the general combination operator $\otimes$, resulting in $f$; remove all the deputing functions and replace each occurrence of a deputy variable with the corresponding bastard variable; multiply $f$ together with all the normal factors; output the resulting faction; and exit.

2. Remove the first variable $z$ from the ordering $\rho$.

3. Remove from $\mathcal{F}$ all the heterogeneous factors $f_1, \ldots, f_k$ that involve $z$, and set

$$f =_{def} \otimes_{i=1}^{k} f_i.$$

Let $B$ be the set of all the variables that appear in $f$.

4. Remove from $\mathcal{F}$, all the normal factors $g_1, \ldots, g_m$ that involve $z$, and set

$$g =_{def} \prod_{j=1}^{m} g_j.$$

Let $C$ be the set of all the variables that appear in $g$.

5. **If** $k=0$, define a function $h$ by

$$h(C-\{z\})=_{def}\sum_z g(C),$$

Add $h$ into $\mathcal{F}$ as a normal factor,

6. **Else if** $m=0$, define a function $h$ by

$$h(B-\{z\})=_{def}\sum_z f(B),$$

Add $h$ into $\mathcal{F}$ as a heterogeneous factor,

7. **Else** define a function $h$ by

$$h(B\cup C-\{z\})=_{def}\sum_z f(B)g(C),$$

Add $h$ into $\mathcal{F}$ as a heterogeneous factor.
**Endif**

8. Recursively call PROJECTION($\mathcal{F}, A, \rho$)

The correctness of PROJECTION is guaranteed by Propositions 4, 5, and 6.

Note that in the algorithm *summing out a variable requires combining only the factors that involve the variable*. This is why PROJECTION allows one to exploit intercausal independencies for efficiency gains. If one ignores intercausal independencies, to sum out one variable one needs combine all the conditional probabilities that involve the variable. There is a gain in efficiency by using PROJECTION because intercausal independence allows one to further factorize conditional probabilities into factors that involve less variables. In Figure 1, for instance, summing out $a$ requires combining $P(e_1|a,b)$ and $P(e_2|a,b,c)$ when intercausal independencies are ignored; there are five variables involved here. By using PROJECTION, one needs to combine $f_{11}(e_1,a)$ and $f_{21}(e_2,a)$; there are only three variables involved in the case.

Finally, we would like to remark that the algorithm is an extension to a simple algorithm for computing marginal probabilities from a homogeneous factorization (Zhang and Poole 1994).

## 9  An example

To illustrate PROJECTION, consider computing the conditional probability $P(e_2|y=0)$ in the Bayesian network $\mathcal{N}$ shown in Figure 2. Since $P(e_2|y=0)$ can be readily obtained from the marginal probability $P(e_2, y)$, we shall show how PROJECTION computes the latter.

Suppose the elimination ordering $\rho$ is: $e_3$, $e_3'$, $a$, $b$, $e_1$, $e_1'$, $c$. Initially, the factors are as follows:

- $\mathcal{F}_0 = \{f_{11}(e_1,a), f_{12}(e_1,b), f_{21}(e_2,a), f_{22}(e_2,b), f_{23}(e_2,c), f_{31}(e_3,e_1'), f_{32}(e_3,e_2')\};$
- $\mathcal{F}_1 = \{P(a), P(b), P(c), P(y|e_3'), I_1(e_1,e_1'), I_2(e_2,e_2'), I_3(e_3,e_3')\}.$

The bastard variable $e_3$ appears in heterogeneous factors $f_{31}(e_3 e_1')$ and $f_{32}(e_3, e_2')$, and in the normal factor $I_3(e_3, e_3')$. After summing out $e_3$ the factors become:

- $\mathcal{F}_0 = \{\psi_1(e_1', e_2', e_3'), f_{11}(e_1,a), f_{12}(e_1,b), f_{21}(e_2,a), f_{22}(e_2,b), f_{23}(e_2,c)\};$
- $\mathcal{F}_1 = \{P(a), P(b), P(c), P(y|e_3'), I_1(e_1,e_1'), I_2(e_2,e_2')\},$

where

$$\psi_1(e_1',e_2',e_3')=_{def}\sum_{e_3}(f_{31}(e_3,e_1')\otimes f_{32}(e_3,e_2'))I_3(e_3,e_3').$$

Now $e_3'$ is the next to sum out. $e_3'$ appears in the heterogeneous factor $\psi_1$ and the normal factor $P(y|e_3')$. After summing out $e_3'$, the factors become:

- $\mathcal{F}_0=\{\psi_2(e_1',e_2',y), f_{11}(e_1,a), f_{12}(e_1,b), f_{21}(e_2,a), f_{22}(e_2,b), f_{23}(e_2,c)\};$
- $\mathcal{F}_1 = \{P(a), P(b), P(c), I_1(e_1,e_1'), I_2(e_2,e_2')\},$

where

$$\psi_2(e_1',e_2',y)=_{def}\sum_{e_3'}\psi_1(e_1',e_2',e_3')P(y|e_3').$$

Next, summing out $a$ gives us:

- $\mathcal{F}_0=\{\psi_3(e_1,e_2), \psi_2(e_1',e_2',y), f_{12}(e_1,b), f_{22}(e_2,b), f_{23}(e_2,c)\};$
- $\mathcal{F}_1=\{P(b), P(c), I_1(e_1,e_1'), I_2(e_2,e_2')\},$

where

$$\psi_3(e_1,e_2) =_{def} \sum_a P(a)[f_{11}(e_1,a)\otimes f_{21}(e_2,a)]$$
$$= \sum_a P(a)f_{11}(e_1,a)f_{21}(e_2,a).$$

Then, summing out $b$ gives us:

- $\mathcal{F}_0=\{\psi_4(e_1,e_2), \psi_3(e_1,e_2), \psi_2(e_1',e_2',y), f_{23}(e_2,c)\};$
- $\mathcal{F}_1= \{P(c), I_1(e_1,e_1'), I_2(e_2,e_2')\},$

where

$$\psi_4(e_1,e_2) =_{def} \sum_b P(b)[f_{12}(e_1,b)\otimes f_{22}(e_2,b)]$$
$$= \sum_b P(b)f_{12}(e_1,b)f_{22}(e_2,b).$$

The next variable on $\rho$ is $e_1$, which appears in heterogeneous factors $\psi_3(e_1,e_2)$ and $\psi_4(e_1,e_2)$ and normal factor $I_1(e_1,e_1')$. After summing out $e_1$ the factors become:

- $\mathcal{F}_0=\{\psi_5(e_2,e_1'), \psi_2(e_1',e_2',y), f_{23}(e_2,c)\};$

- $\mathcal{F}_1 = \{P(c), I_2(e_2, e'_2)\}$,

where

$$\psi_5(e_2, e'_1) =_{def} \sum_{e_1} I_1(e_1, e'_1)[\psi_3(e_1, e_2) \otimes \psi_4(e_1, e_2)].$$

Due to space limit, we have to discontinue the example here. Hopefully, the following two points shoul be be clear now. First, in summig out one variable, PROJECTION combines only the factors that involve the variable.

Second, since $e_1$ is a bastard variable, we usually do *not* have

$$\sum_{e'_1} \psi_2(e'_1, e'_2, y)[\sum_{e_1} I_1(e_1, e'_1)(\psi_4(e_1, e_2) \otimes \psi_3(e_1, e_2))] = \sum_{e_1}[\sum_{e'_1} \psi_2(e'_1, e'_2, y) I_1(e_1, e'_1)] \otimes \psi_4(e_1, e_2) \otimes \psi_3(e_1, e_2).$$

This is why $e'_1$ must be summed out *after* $e_1$ has been sumed out.

## 10 RELATED WORK

The paper is closely related to Heckerman (1993). However, the relationship between Heckerman's temporal definition of intercausal independence and our constructive definition remains to be clarified.

Our constructive definition is a special case of the generalization of the noisy-OR model proposed by Srinivas (1993). While we consider only binary operators for combining contributions from different sources, Srinivas considers general mappings from contributions to effect.

In the additive belief-network model proposed by Dagum and Galpher (1993), a conditional probability $P(y|x_1, \ldots, x_n)$ is decomposed into a linear combination of the $P(y|x_i)$'s. Although this may appear to be a special case of constructive equation (3), they are actually very different. For example, Lemma 2 would not hold if $\otimes$ were addition.

## 11 CONCLUSION

A constructive definition of intercausal independence has been given. The definition is based one an intuitive picture of intercausal independence where a number of causes contribute independently to an effect and the total contribution is a combination of the individual contributions. Noisy OR-gates and noisy adders are examples of constructive intercausal independence.

It is well known that conditional independence implies factorization of joint probability, which enables one to make use of conditional independencies to reduce inference complexity in Bayesian networks. Under the constructive definition, intercausal independence implies factorization of conditional probability, which allows one to make use of intercausal independencies, together with conditional independencies, to further reduce inference complexity.

### Acknowledgement

The authors are grateful for the three anonymous reviewers for their valuable comments and suggestions. Research is supported by NSERC Grant OGP0044121 and by a travel grant from Hong Kong University of Science and Technology.